\documentclass[11pt]{extarticle}
\usepackage{graphicx}
\usepackage{microtype} 
\usepackage{booktabs} 
\usepackage{tabularx}
\usepackage{caption}
\usepackage{amsmath}
\usepackage{amssymb}
\usepackage{float}
\usepackage{multirow}
\usepackage{xcolor}
\usepackage{algorithm}
\usepackage{algpseudocode}
\usepackage{url}
\usepackage{doi}
\usepackage[toc,page]{appendix}
\usepackage[margin=2.8cm]{geometry}
\usepackage{titlesec}
\usepackage{hyperref}

\begin{document}

\title{Decision Transformer vs. Decision Mamba: Analysing the Complexity of Sequential Decision Making in Atari Games}
\author{Ke Yan\\The University of Edinburgh\\k.yan-9@sms.ed.ac.uk}
\date{}
\maketitle

\begin{abstract} 
This work analyses the disparity in performance between Decision Transformer (DT) and Decision Mamba (DM) in sequence modelling reinforcement learning tasks for different Atari games. The study first observed that DM generally outperformed DT in the games Breakout and Qbert, while DT performed better in more complicated games, such as Hero and Kung Fu Master. To understand these differences, we expanded the number of games to 12 and performed a comprehensive analysis of game characteristics, including action space complexity, visual complexity, average trajectory length, and average steps to the first non-zero reward. In order to further analyse the key factors that impact the disparity in performance between DT and DM, we employ various approaches, including quantifying visual complexity, random forest regression, correlation analysis, and action space simplification strategies. The results indicate that the performance gap between DT and DM is affected by the complex interaction of multiple factors, with the complexity of the action space and visual complexity (particularly evaluated by compression ratio) being the primary determining factors. DM performs well in environments with simple action and visual elements, while DT shows an advantage in games with higher action and visual complexity. Our findings contribute to a deeper understanding of how the game characteristics affect the performance difference in sequential modelling reinforcement learning, potentially guiding the development of future model design and applications for diverse and complex environments.
\end{abstract}

\let\thefootnote\relax\footnotetext{Project code available: \href{https://github.com/0x1DA9430/Decision-Transformer-vs-Decision-Mamba}{Github}}

\section{Introduction}
The field of reinforcement learning (RL) has seen significant advancements in recent years, particularly in the domain of sequential decision-making tasks. One notable approach has emerged as a powerful tool for these challenges: the Decision Transformer (DT) \cite{chen2021decision}. Recently, Ota introduced Decision Mamba (DM), which replaces DT's causal self-attention blocks with the new Mamba architecture \cite{ota2024decision, gu2023mamba}. Both DT and DM have shown promise in various RL tasks. However, to the best of our knowledge, a comprehensive analysis of their relative performance across different environments remains an open question. This study aims to fill this gap by thoroughly comparing DT and DM across a wide range of Atari games.

Our investigation begins with an initial observation: while DM outperforms DT in some games, such as Breakout, DT demonstrates superior performance in others, notably Hero and KungFuMaster. This gap in performance across different games raises questions about the factors that influence the effectiveness of these models. We seek to understand whether certain architectural features of DT and DM are better suited to specific types of games, and how various game characteristics impact their relative performance.

Based on previous observations, we perform a sequence of analyses that integrate empirical evaluation with an in-depth study of game characteristics. This investigation begins with a comprehensive survey of various game characteristics, including action-space complexity and visual complexity. To quantify the visual complexity, we introduce three metrics: image entropy, compression ratio, and feature count. These metrics, along with other game characteristics such as the number of actions and average trajectory length, form the basis for detailed analysis using random forest regression and correlation studies. The analysis results show that the action space complexity is indeed an important factor in determining the performance of the model, but at the same time, the visual complexity also plays a crucial role. Finally, to isolate the effect of action space complexity and explore further, two action space simplification strategies are implemented and evaluated: simple action fusion and frequency-based action fusion. These studies provide more insight into the influence of action space complexity and expose the limitations of simplification strategies in maintaining game dynamics.

The key contributions of this study are:
\begin{enumerate}
    \item A thorough comparison of Decision Transformer (DT) and Decision Mamba (DM) in a variety of Atari games, providing insights into their relative strengths and weaknesses.
    \item The identification and analysis of key game characteristics that influence model performance, with a focus on visual and structural elements of the games.
    \item A quantitative assessment of the relative importance of various game characteristics in determining the performance gap between DT and DM, using Random Forest regression and correlation analysis.
    \item The development and evaluation of novel strategies to simplify the action space in complex games.
\end{enumerate}
Our findings reveal that while action space complexity is indeed a significant factor in determining model performance, visual complexity, particularly as measured by compression ratio, also plays a crucial role. We observe that DM tends to excel in visually simpler environments, while DT shows advantages in games with higher visual complexity. These insights not only contribute to our understanding of the strengths and limitations of DT and DM but also have implications for the design and application of sequence modelling approaches in RL.

The rest of the dissertation is structured as follows: Section 2 provides background information on Transformer, State Space Models, basic concepts of Reinforcement Learning, and related work on applying these sequence modelling architectures to reinforcement learning tasks. Section 3 presents our approach in detail, including dataset preparation, model architecture, and experimental setup. Section 4 focuses on our experiments and analysis. It begins with a basic comparison and extended measurements of the game's action space complexity and visual complexity. This is followed by the application of random forest, correlation analysis, and action fusion to further explore the effect of action complexity. Section 5 discusses potential directions for future work, and Section 6 summarizes the findings of this paper. Through these analyses, this study aims to provide valuable insights into the factors that affect the performance of sequence modelling in reinforcement learning tasks, possibly guiding future development of model design and applications in more complex environments.

\section{Background}
\subsection{Transformer}
The Transformer introduced by Vaswani et al. revolutionized the field of sequence modelling \cite{vaswani2017attention}. The architecture consists of an encoder and decoder, each of which contains multiple layers of self-attention and fully connected neural networks, with a cross-attention connecting the encoder and decoder  \cite{vaswani2017attention}. Compared to Recurrent Neural Networks (RNNS), Transformer can process the entire sequence simultaneously, and without suffering long-range dependency problems. This is achieved through the self-attention mechanism. It allows the model to weigh the importance of different parts of the input sequence, captures complex relationships inside the data, and can be computed in parallel on modern GPUs. The Transformer show better performance on various tasks while significantly reducing training time \cite{vaswani2017attention, devlin2018bert}. However, they may encounter difficulties when dealing with very long sequences during inference due to their quadratic computational complexity, which leads to a large video RAM cost. \cite{gu2023mamba}.

\subsection{State Space Models}
State Space Models (SSMs) have emerged as a promising alternative for long sequential modelling tasks. An SSM, or state space model, utilises a mathematical framework to convert a one-dimensional input into an n-dimensional latent state, which is then projected back to a one-dimensional output \cite{gu2023mamba, gu2022s4}. The process can be expressed by the following equations:
\begin{align}
    h'(t) &= Ah(t) + Bx(t) \label{eq:h_prime} \\
    y(t) &= Ch(t) \label{eq:y}
\end{align}
where $h(t)$ represents the $N$-D latent state, $x(t)$ is the $1$-D input, and $y(t)$ is the $1$-D output \cite{gu2023mamba, gu2022s4}.

The equations of SSMs are similar to those of RNNs, but there are differences in how the matrix $A$ is initialised. In traditional recurrent neural networks (RNNs), these matrices are commonly initialised randomly and then updated via backpropagation. In contrast, SSMs utilise HiPPO (High-order Polynomial Projection Operators) theory for initialization. The HiPPO theory initializes the A matrix to represent the projection of the input sequence onto polynomial bases \cite{gu2020hippo}. For instance, one variant of the HiPPO matrix, known as HiPPO-LegT:
\begin{equation}
    A_{nk} = -\begin{cases}
\left(2n + 1\right)^{1/2}\left(2k + 1\right)^{1/2} & \text{if } n > k \\
n + 1 & \text{if } n = k \\
0 & \text{if } n < k .
\end{cases}
\end{equation}
where, $n$ and $k$ are indices representing the row and column of the matrix $A$, respectively \cite{gu2020hippo}.

This initialization allows SSMs to efficiently capture information across different temporal scales, hence enhancing the modelling of long-range dependencies in comparison to RNNs \cite{gu2022s4,gu2020hippo}.

\subsubsection{Mamba}
Despite the benefits of structured HiPPO initialization, there are limitations to the original SSM due to its time-invariant nature, which means that the matrices $A$, $B$, and $C$ are constant throughout time. The Mamba architecture, introduced by Gu et al., addresses this problem by incorporating a selection mechanism that introduces linear time-varying \cite{gu2023mamba}. Inspired by gating mechanisms in Long Short-Term Memory (LSTM) networks, which have been successful in selectively updating and forgetting information in sequential data processing \cite{hochreiter1997lstm}. In Mamba, this selection mechanism allows the model to adaptively focus on or ignore specific inputs based on their relevance, making it capture long-range dependencies and selectively copy from the inputs \cite{gu2023mamba}.

\begin{figure}[H]
    \centering
    \includegraphics[width=0.39\linewidth]{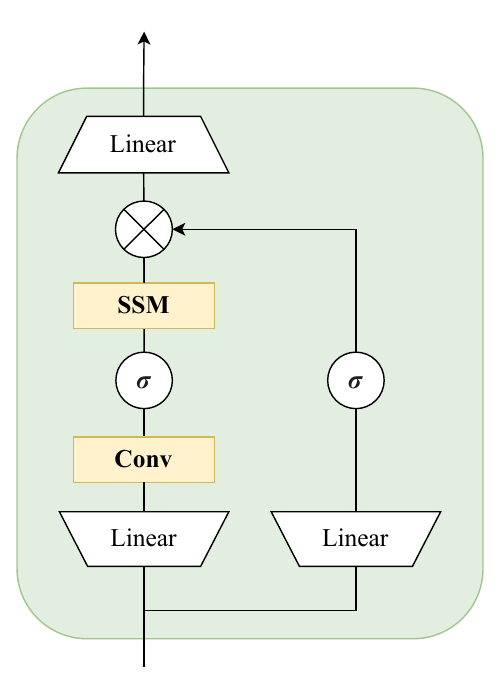}
    \caption{Mamba block, $\sigma$ is SiLU function, and $\otimes$ stands for elementwise multiplication.}
    \label{fig:mamba-block}
\end{figure}

As illustrated in Figure \ref{fig:mamba-block}, the core component of the Mamba architecture is the Mamba block. The main branch starts with a linear layer, succeeded by a convolutional layer. This output then goes through SiLU (Sigmoid Linear Unit) activation before being processed by the SSM. The second branch includes a linear layer and SiLU activation. The outputs of both branches are combined using element-wise multiplication and then pass through a final linear layer. This architecture enables Mamba to efficiently model complex dependencies and adapt its processing based on input content. The Mamba architecture has demonstrated promising results across various domains, including language modelling, DNA sequence modelling, and audio waveform processing \cite{gu2023mamba}. Its ability to handle long sequences efficiently while maintaining selective copying capabilities makes it a competitive alternative to traditional Transformer models in certain applications.

\subsection{Reinforcement Learning}
\subsubsection{Traditional Reinforcement Learning}
Reinforcement learning (RL) is a paradigm in machine learning. In traditional RL, the agent (model) learns to make decisions by interacting with the environment. One of the basic frameworks of RL is the Markov decision process (MDP), which assumes that the future state of the environment depends only on the current state and behaviour, not on the previously recorded history (Markov property)\cite{sutton2018reinforce}. For example, Q-learning is a common method in MDP. It is a value-based method that uses a Q-function to estimate the expected cumulative reward for each state-action pair, and then produce policy for operation actions. A simplified representation of the Q-function can be expressed as:
\begin{equation}
    Q(s, a) = r + \gamma \cdot \max{Q(s', a')}
\end{equation}
where $s$ is the current state, $a$ is the current action, $r$ is the immediate reward, $s'$ is the next state, $a'$ represents all possible actions in the next state, and $\gamma$ is the discount factor for future rewards \cite{sutton2018reinforce}. This equation, also known as the Bellman equation, uses Markov properties by considering only the immediate reward and the value of the next state-action pair, allowing efficient recursive computation. Q-learning has achieved significant success in many applications, from game playing to robot control. However, they often struggle with sample efficiency problems, especially in environments with sparse rewards \cite{sutton2018reinforce}.

\subsubsection{Sequence Modeling in Reinforcement Learning}
In recent years, there has been a paradigm shift in solving reinforcement learning tasks. Motivated by the success of sequence models across several domains, the new method redefines the reinforcement learning task as a sequence modelling problem. The Decision Transformer (DT), introduced by Chen et al., is a prime example of this paradigm shift in reinforcement learning \cite{chen2021decision}. Utilizing the power of the Transformer in sequence modelling, DT processes trajectories of return-to-go, states, and actions to predict the next optimal actions. DT reframes the RL problem as a sequence prediction task, freeing itself from the Markov property assumption by considering the entire trajectory rather than just the immediate state-action pairs. Therefore this approach does not require and explicit value-function (Bellman equation) approximation, which is fundamentally different from the traditional RL. The larger context allows DT to consider longer sequences of states, actions, and returns, enabling it to directly learn the relationship between returns and optimal actions. The key advantage of this approach is that it can consider a wider range of contexts, thus capturing long-term dependencies more effectively than traditional RL methods. Moreover, DT uses offline data for training, avoiding the inefficiency caused by the agent continuously interacting with the environment in traditional RL. 

Decision Transformer represent a promising direction in reinforcement learning research. Based on the Decision Transformer, several variants have been proposed to solve specific challenges or improve performance in different domains. Decision ConvFormer replaces the causal-self attention layer of DT with a convolutional layer. For state, action, and return-to-go in the game trajectory, it applies three separate convolutional filters: The state filter, action filter, and RTGfilter to improve the model's ability to capture the inherent local correlations \cite{kim2023decisionconv}. The Decision S4 model focuses on continuous control tasks such as HalfCheetah. It uses a structured state space sequence (S4) model to improve the efficiency of modelling long-range dependencies. With fewer parameters and training time, it achieves better performance than Decision Transformer on most tasks. \cite{bardavid2023decisions4}. Recently, Ota introduced Decision Mamba (DM), replacing DT's causal self-attention mechanism with the Mamba blocks \cite{ota2024decision}. This modification aims to utilise the efficiency and effectiveness of the Mamba architecture in handling long sequences, showing competitive performance with DT models.

While these studies show promising results, there remains a gap in understanding how different sequence modelling architectures apply to various reinforcement learning tasks. To the best of our knowledge, most existing work has focused on limited environments. For instance, Decision S4 concentrates on continuous control tasks in MuJoCo. Alternatively, research has examined specific aspects of model performance, such as the ability to capture local correlations (Decision ConvFormer) or long-range dependencies (Decision S4 and Decision Mamba), leaving room for a more comprehensive comparative analysis. Moreover, the impact of game characteristics, such as action space complexity and visual complexity, on the performance of these models has not been systematically investigated.

\section{Methodology}
\subsection{Dataset}
This study utilizes the DQN-replay dataset introduced by Agarwal et al., comprising the replay experience of a DQN agent across Atari games \cite{agarwal2020dqn}. This dataset contains approximately 50 million trajectories, each composed of (state, action, reward, next state) tuples. States are represented as 84$\times$84 pixel frame stacks of four consecutive images, providing temporal context for the agent's decision-making process. The action space and reward structure depend on the specific game mechanics.

\subsubsection{Data Processing}
The first step of implementing Decision Transformer and Decision Mamba is the preparation of input data. This process involves transforming raw game experiences into a format suitable for sequence modelling. Basically, we followed the method used in the original Decision Transformer paper \cite{chen2021decision}.

The primary innovation in data representation for these models is the use of return-to-go (RTG) instead of immediate rewards. Return-to-go at any given timestep is defined as the sum of all future rewards from that point until the end of the episode \cite{chen2021decision, ota2024decision}. For a timestep $t$ in an episode of length $T$, the return-to-go $\hat{R}_t$ is calculated as:
\begin{equation}
    \hat{R}_t = \sum_{t'=t}^T r(t')
\end{equation}
where $r(t')$ is the reward at timestep $t'$.

In the data preparation process, each trajectory is represented as a sequence of return-to-go values, states, and actions. The sequence takes the form:
\begin{equation}
    (\hat{R}_1, s_1, a_1, \hat{R}_2, s_2, a_2, ..., \hat{R}_T, s_T, a_T)
\end{equation}
where $s_t$ represents the state at timestep $t$, and $a_t$ is the action taken at that timestep $t$.

With data prepared, the training set is then created by sampling trajectories from the DQN-replay dataset. To reduce computational overhead, we followed the approach of Chen et al., that is, for each game, randomly sampling 1\% of the total gameplay experience \cite{chen2021decision}. Consequently, the model is trained offline using this pre-collected gameplay experience, allowing for efficient training without requiring real-time interaction with the game environment. Following the offline training, we conduct the online evaluation. During evaluation, we do not use a separate test set from the dataset. Instead, we evaluate the trained model directly in the Atari game environment using the Arcade Learning Environment (ALE), which is specifically focused on Atari games and now serves as a foundational backend for OpenAI Gym \cite{bellemare13arcade}. This approach consistent with the standard practice in reinforcement learning, ensures that the model's performance is assessed on unseen game states and trajectories, effectively preventing any potential information leakage from the training data to the evaluation phase.

\subsection{Model Architecture}

\begin{figure}[h!]
    \centering
    \includegraphics[width=0.69\linewidth]{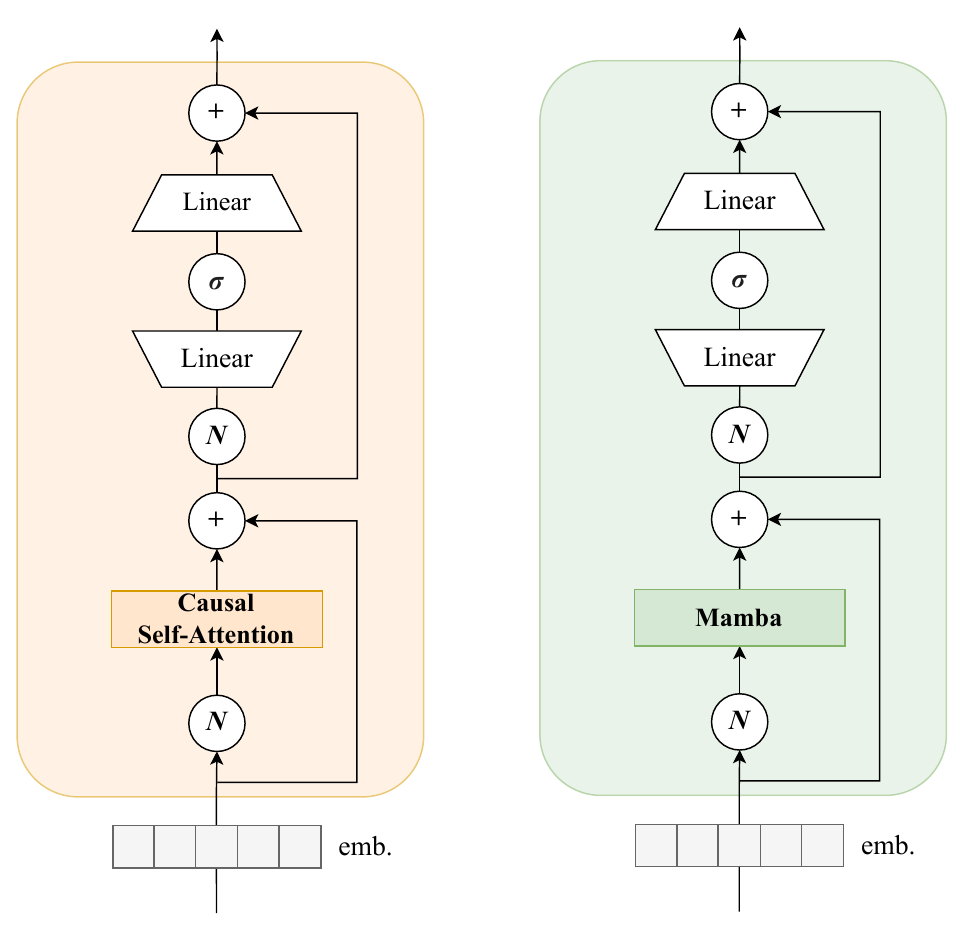}
    \caption{The Architecture of Decision Transformer (Left) and Decision Mamba (Right). \(N\) represents normalization layers, activation function \(\sigma\) stands for GELU (Gaussian Error Linear Unit), and \(+\) are addition operations used for skip connections.}
    \label{fig:dt-dm-arc}
\end{figure}

The architecture of the Decision Transformer, as illustrated in Figure \ref{fig:dt-dm-arc} (left), consists of several key components. The input to the model is a sequence of tokens representing returns-to-go, states, and actions. These inputs are first processed through embedding layers and combined with positional embeddings to provide temporal context. The core of the model is a stack of transformer layers, each containing a causal self-attention mechanism and a feedforward neural network. This structure allows the model to capture complex dependencies between different elements of the input sequence \cite{chen2021decision}.

Decision Mamba builds on the Decision Transformer by replacing the causal self-attention mechanism with the Mamba block. As shown in Figure \ref{fig:dt-dm-arc} (right), Decision Mamba's overall structure is similar to the Decision Transformer, with the key difference being the replacement of the attention mechanism \cite{ota2024decision}.

Both DT and DM operate on similar principles when it comes to training and inference. During training, these models learn to predict the next action given the current state, past actions, and the desired return-to-go. At inference time, both models generate actions autoregressively. The process begins by specifying an expected return. The model then observes the current state and predicts the next action, which is executed in the environment to obtain a new state and reward. The return-to-go is updated by subtracting the received reward, and this process continues until the episode terminates.

\subsection{Experimental Setup}
Our experimental design aimed to evaluate and compare the performance of Decision Transformer (DT) and Decision Mamba (DM) across various Atari games. We initially selected four games: Breakout, Qbert, Hero, and KungFuMaster. Breakout and Qbert were chosen as they were previously examined in both the Decision Transformer and Decision Mamba papers, providing a baseline for comparison. Hero and KungFuMaster were added to expand the scope of our analysis, as they potentially present greater challenges for the models due to their more complex action spaces and game dynamics. 

To ensure a comprehensive evaluation, we conducted experiments with different context lengths: 10, 30, and 50, where 30 is the default value in the original Decision Transformer paper. For all experiments, we maintained consistency in hyperparameters across both models, the detailed settings of all hyperparameters can be found in Appendix \ref{sec:hyper}.

In the extended experiments, we broadened our game selection to include eight additional Atari games, bringing the total to twelve, which aims to provide a more complete view of how DT and DM perform across a wider range of game characteristics and complexities. For these additional experiments, we focused on the context length of 10 for computational efficiency while still providing valuable insights.

\subsubsection{Loss function}
Atari games, which are considered discrete action control problems, differ from continuous control tasks like those in robotics. The Atari games have a fixed set of possible actions such as UP, Down, Left, Right, and Fire. For these discrete action predictions, the cross-entropy loss is particularly well-suited and widely used in sequence modelling \cite{chen2021decision}. The equation of cross-entropy loss can be defined as:
\begin{equation}
    L = -\sum_{a=0}^{M} y_{s,a} \log(p_{s,a})
\end{equation}
where \( M \) is the number of actions, \( y \) is 1 or 0, depends on if action label \( a \) is correct or not for state \( s \), and \( p \) is the predicted probability of action \( a \) at state \( s \).

\subsubsection{Evaluation}
The model is trained offline using the DQN-replay dataset and subsequently assessed online using the Arcade Learning Environment (ALE) \cite{bellemare13arcade}. This methodology accelerates the training process by avoiding continuous interaction with the game environment. Furthermore, in the original paper of Decision Transformer, it has been shown to achieve results comparable to online learning \cite{chen2021decision}.

\paragraph{Expected Return Setup}
As outlined in the original Decision Transformer paper, the expected return acts as a target for the model to aim for during the process of inference. The value is initialised at the start of every episode and serves as a guiding factor for the model's decision-making process \cite{chen2021decision}.

The original DT paper used different multipliers when setting the expected return for each game. For Breakout, they used 1 times the maximum return in the dataset, while for Qbert, they used 5 times the maximum. In our study, we aimed for consistency across all games. Therefore, we chose to use 5 times the maximum return observed in the DQN-play dataset for all games. This is also based on a naive assumption, that is, setting a relatively optimistic initial value can be beneficial for the agent's performance \cite{sutton2018reinforce}. This assumption has been validated in conventional RL methods but remains an open question to these novel architectures such as DT and DM. It could be an interesting future research direction to explore the impact of different initial expected returns. In this article, we will only initialize it as 5 times the maximum return occurred in the dataset.

\begin{table}[h!]
\centering
\small
\setlength{\tabcolsep}{15pt}
\begin{tabular}{@{}lcc@{}}
\toprule
\textbf{Games} & \textbf{Max Return} & \textbf{Expected Return (5*max)} \\ 
\midrule
Breakout       & 104          & 520                 \\ 
Qbert          & 640          & 3200                \\ 
Hero           & 190          & 950                 \\ 
KungFuMaster   & 284          & 1420                \\ 
\bottomrule
\end{tabular}
\captionsetup{width=0.75\linewidth}
\caption{Max Return Observed in Dataset, and Expected Return Setup for Evaluation (Excerpt; full table in Appendix).}
\label{tab:exp-return}
\end{table}

Table \ref{tab:exp-return} illustrates our setup for four games: Breakout, Qbert, Hero, and KungFuMaster. For example, in Breakout, where the maximum observed reward in our training set was 104, we set the expected return to 520. During evaluation, we apply the same expected return for both DT and DM.

\paragraph{Normalized Score}
To account for the scoring mechanisms across different games, we employ a normalized score metric. This approach, following the methodology of Hafner et al. and Ye et al., allows for meaningful comparisons of performance across different game environments \cite{hafner2021master, ye2021master}. This normalization maps the performance onto a scale where 0 represents a random walk score and 100 represents human-level performance. The normalized score is calculated using the following equation:

\begin{equation}
    \textit{SCORE}_{\textit{normalized}} := 100 \times \frac{\textit{SCORE}_{\textit{raw}} - \textit{SCORE}_{\textit{random}}}{\textit{SCORE}_{\textit{human}} - \textit{SCORE}_{\textit{random}}}
\end{equation}\\
where: $\textit{SCORE}_{\textit{raw}}$ is the actual score achieved by the agent (or model).

For $\textit{SCORE}_{\textit{random}}$ and $\textit{SCORE}_{\textit{human}}$, we follow the results reported by Hafner et al. and Ye et al., where the random scores were obtained by agents taking random actions and human scores were collected from players given 2 hours to practice each game \cite{hafner2021master, ye2021master}. Table \ref{tab:rand-human-score} presents the random and human benchmark scores:

\begin{table}[H]
\centering
\small
\begin{tabular}{ccc}
\toprule
\textbf{Game} & \textbf{Random Walk} & \textbf{Human Players} \\
\midrule
Breakout & 1.7 & 30.5 \\
Qbert    & 163.9 & 13455.0 \\
Hero     & 1027.0 & 30826.4 \\
KungFuMaster & 258.5 & 22736.3 \\
\bottomrule
\end{tabular}
\captionsetup{width=0.8\linewidth}
\caption{Game scores for Random Walk and Human Player (Excerpt; full table in Appendix \ref{sec:scores}).}
\label{tab:rand-human-score}
\end{table}

This normalization allows us to assess the relative performance of our models across different games, regardless of their varying score scales. The full table of random walk and human benchmark scores for all games in our study can be found in Appendix \ref{sec:scores}.

\section{Experiments and Analysis}

\subsection{Initial Experiments}
\begin{table}[h!]
\centering
\small      
\begin{tabularx}{\textwidth}{@{}X *{6}{>{\raggedleft\arraybackslash}X}@{}}
\toprule
               & \multicolumn{3}{c}{Decision Transformer (DT)} & \multicolumn{3}{c}{Decision Mamba (DM)} \\
               \cmidrule(lr){2-4} \cmidrule(lr){5-7}
Game           & 10                    & 30                    & 50                    & 10                    & 30                    & 50                    \\ \midrule
Breakout       & 288.11    & 251.39    & 238.19    & \textbf{343.13}    & \textbf{384.79}    & \textbf{401.39}   \\[-1ex]
               & $\pm$ 89.98    & $\pm$ 74.14    & $\pm$ 41.33    & \textbf{$\pm$ 65.35}    & \textbf{$\pm$ 54.29}    & \textbf{$\pm$ 59.44}   \\
\midrule
Qbert          & \textbf{26.73}     & 4.94     & 11.48      & 26.54      & \textbf{26.10}     & \textbf{22.59}      \\[-1ex]
               & \textbf{$\pm$ 7.74}     & $\pm$ 2.09     & $\pm$ 6.12      & $\pm$ 1.01      & \textbf{$\pm$ 4.20}     & \textbf{$\pm$ 1.84}      \\
\midrule
Hero           & \textbf{31.63}      & \textbf{28.05}      & \textbf{27.90}      & 7.70       & 6.95       & 7.69       \\[-1ex]
               & \textbf{$\pm$ 4.16}      & \textbf{$\pm$ 4.44}      & \textbf{$\pm$ 7.41}      & $\pm$ 0.96       & $\pm$ 0.43       & $\pm$ 1.42       \\
\midrule
KungFuMaster   & \textbf{29.41}      & \textbf{14.23}      & 6.98     & 6.32       & 7.42       & \textbf{7.23}       \\[-1ex]
               & \textbf{$\pm$ 6.47}      & \textbf{$\pm$ 5.89}      & $\pm$ 2.35     & $\pm$ 1.75       & $\pm$ 1.46       & \textbf{$\pm$ 0.98}       \\
\bottomrule
\end{tabularx}
\caption{Normalized score (mean $\pm$ std) of Decision Mamba and Decision Transformer across four games with context lengths of 10, 30, and 50. Each game was run with 5 random seeds per context length (outliers removed to reduce standard deviation while retaining at least 3 data points; the raw normalized scores, without outlier removal, are provided in Appendix \ref{sec:raw-norm-score}). \textbf{Bold} indicates the better mean performance.}
\label{tab:find-the-gap}
\end{table}

The experimental design aims to compare the performance of Decision Transformer (DT) and Decision Mamba (DM) across four Atari games: Breakout, Qbert, Hero, and Kung Fu Master \footnote{Breakout and Qbert were examined in the original DT and DM papers, and we reproduced these results. Hero and Kung Fu Master were selected by us to expand the scope of the analysis.}. These games were chosen to represent a range of complexities and challenges, allowing for an evaluation of the models' capabilities.

Our experiments also explored the effect of different context lengths (10, 30, and 50), with 30 being the default setting in the original Decision Transformer paper. Table \ref{tab:find-the-gap} presents the normalized scores for both models across the four selected games. The results reveal several notable patterns:
\begin{enumerate}
    \item DM consistently outperforms DT in Breakout across all context lengths. The performance gap is especially significant at context length 50.
    \item Performance in Qbert is mixed, with DT performing slightly better at context length 10, but with high standard deviation. DM shows a great advantage when context lengths go to 30 and 50.
    \item DT significantly outperforms DM in Hero across all context lengths. The performance gap is significant, with DT achieving scores around 30, while DM's scores remain below 8.
    \item In Kung Fu Master, DT shows a clear advantage at context lengths 10 and 30, but its performance significantly drops as the context length increases.
\end{enumerate}

To understand the significant performance differences, particularly in Hero and KungFuMaster, we quantify the complexity of each game based on various metrics derived from the DQN-replay dataset. The analysis is shown in Table \ref{tab:game-analysis}.

\begin{table}[H]
\centering
\small
\begin{tabular}{|l|c|c|c|}
\hline
\textbf{Game} & \textbf{\# Actions} & \textbf{Avg. Trajectory} & \textbf{Avg. Steps to} \\
 & & \textbf{Length} & \textbf{First non-zero Reward} \\
\hline
Breakout & 4 & 1299.62 & 45.20 \\
Qbert & 6 & 1060.84 & 56.75 \\
Hero & 18 & 1192.23 & 54.94 \\
KungFuMaster & 14 & 2642.71 & 109.53 \\
\hline
\end{tabular}
\caption{Analysis results for various games based on DQN-replay dataset.}
\label{tab:game-analysis}
\end{table}

The results suggest that action space complexity significantly influences model performance. Games like Hero and KungFuMaster, with 18 and 14 possible actions respectively, have a large action space compared to Breakout and Qbert, which have only 4 and 6 actions. This difference in the number of unique actions appears to be a key factor in explaining the performance difference between the models.

\subsection{Extended Experiments and Analysis}

\begin{table}[h!]
\centering
\small
\setlength{\tabcolsep}{10pt}
\begin{tabular}{@{}lcc@{}}
\toprule
\textbf{Game} & \textbf{Decision Transformer (DT)} & \textbf{Decision Mamba (DM)} \\
\midrule
Breakout & 309.14 $\pm$ 97.41 & \textbf{367.13 $\pm$ 75.09} \\
Qbert & \textbf{36.98 $\pm$ 11.38} & 26.93 $\pm$ 1.20 \\
Hero & \textbf{30.37 $\pm$ 4.47} & 7.77 $\pm$ 0.99 \\
KungFuMaster & \textbf{29.41 $\pm$ 6.48} & 5.29 $\pm$ 0.89 \\
Pong & \textbf{71.58 $\pm$ 26.82} & 64.31 $\pm$ 53.24 \\
Seaquest & 2.05 $\pm$ 0.43 & \textbf{2.77 $\pm$ 0.35} \\
Alien & \textbf{12.42 $\pm$ 1.51} & 11.74 $\pm$ 2.16 \\
BankHeist & \textbf{0.63 $\pm$ 0.51} & -0.09 $\pm$ 0.56 \\
BattleZone & \textbf{9.49 $\pm$ 5.39} & 7.20 $\pm$ 1.58 \\
RoadRunner & 25.00 $\pm$ 8.19 & \textbf{28.49 $\pm$ 8.00} \\
FishingDerby & \textbf{160.44 $\pm$ 28.23} & 153.14 $\pm$ 6.43 \\
SpaceInvaders & 26.06 $\pm$ 1.88 & \textbf{28.77 $\pm$ 2.95} \\
\bottomrule
\end{tabular}
\captionsetup{width=0.913\linewidth}
\caption{Normalized score (mean $\pm$ std) of Decision Mamba and Decision Transformer across different games, context length 10. Scores are averaged over 3 separate runs, each with a unique random seed. \textbf{Bold} indicates better scores.}
\label{tab:full-performance-comparison}
\end{table}

While the initial experiments revealed interesting performance differences between Decision Transformer (DT) and Decision Mamba (DM) across four Atari games, we recognized the need for a more comprehensive analysis to understand the factors influencing these disparities. Therefore, we expanded our analysis to a broader range of Atari games. We added eight new games, carefully selected to represent a variety of action space complexities. This expanded dataset allows us to examine how the performance difference between DT and DM varies across a more diverse set of game environments. To maintain computational efficiency, we conducted these additional experiments using a context length of 10.

Table \ref{tab:full-performance-comparison} presents a thorough comparison between Decision Transformer (DT) and Decision Mamba (DM) across 12 Atari games. DT outperforms in 8 of the 12 games, including Qbert, Hero, KungFuMaster, Pong, Alien, BankHeist, BattleZone, and FishingDerby. As reported before, DT's advantage is significant in Hero and KungFuMaster with scores of 30.37 and 29.41 respectively compared to DM's 7.77 and 5.29. Conversely, DM excels in 4 games: Breakout, Seaquest, RoadRunner, and SpaceInvaders.

\begin{table}[h!]
\centering
\small
\begin{tabular}{|l|c|c|c|c|c|c|c|}
\hline
Game & \# Act. & Avg. Traj. & Avg. Steps & Image & Compression & Feature \\
 & & Length & 1st Reward & Entropy & Ratio & Count \\
\hline
Breakout & 4 & 1299.62 & 45.20 & 1.50 & 21.35 & 23.33 \\
Qbert & 6 & 1060.84 & 56.75 & 1.89 & 5.80 & 84.64 \\
Hero & 18 & 1192.23 & 54.94 & 2.01 & 10.55 & 38.84 \\
KungFuMaster & 14 & 2642.71 & 109.53 & 2.66 & 7.58 & 52.63 \\
Pong & 6 & 2096.52 & 112.64 & 0.68 & 40.00 & 9.16 \\
Seaquest & 18 & 1413.12 & 87.23 & 2.24 & 12.92 & 16.18 \\
Alien & 18 & 932.20 & 22.49 & 2.02 & 7.71 & 22.88 \\
BankHeist & 18 & 1185.34 & 20.04 & 1.88 & 7.78 & 188.87 \\
BattleZone & 18 & 2068.26 & 267.60 & 2.84 & 9.31 & 13.88 \\
RoadRunner & 18 & 1123.01 & 81.03 & 1.77 & 12.89 & 24.61 \\
FishingDerby & 18 & 1775.02 & 44.58 & 2.20 & 8.01 & 23.34 \\
SpaceInvaders & 6 & 1820.79 & 52.07 & 0.84 & 13.30 & 65.85 \\
\hline
\end{tabular}
\captionsetup{width=0.913\linewidth}
\caption{Extended analysis results for chosen Atari games based on the DQN-replay dataset. Abbreviations: \# Act. = Number of Actions, Avg. Traj. Len. = Average Trajectory Length, Avg. Steps 1st Reward = Average Steps to First Non-Zero Reward.}
\label{tab:full-game-analysis}
\end{table}

To better understand these performance differences, we conducted an in-depth analysis of various game characteristics. Table \ref{tab:full-game-analysis} presents a set of metrics for each game. We report the number of actions, average trajectory length, and average steps to the first non-zero reward for each game as before. To further capture the visual features of each game, we introduced three additional metrics: Image Entropy, Compression Ratio, and Feature Count:

\begin{itemize}
    \item \textbf{Image Entropy}\\
    Image Entropy is a measure of randomness or unpredictability in the image, by quantifying the amount of information contained in an image \cite{hayashi2023image}. Higher entropy generally indicates more complex, information-rich images. It is calculated using the Shannon entropy formula \cite{shannon1948entropy}:
    \begin{equation}
        H = -\sum{(p_i \cdot \log_2(p_i))}
    \end{equation}
    where $p_i$ is the probability of pixel intensity $i$ occurring in the image. We used OpenCV's calcHist function to compute each frame's pixel intensities histogram. The histogram was then normalized to obtain probability distributions, and Shannon's entropy formula was applied to these distributions. 
    
    \item \textbf{Compression Ratio}\\
    Compression Ratio is the ratio of the uncompressed size to the compressed size of data \cite{salomon2009compress}. Calculated as:
    \begin{equation}
        \textit{Compression Ratio} = \frac{\textit{Uncompressed Size}}{\textit{Compressed Size}}
    \end{equation}
    More complex images are typically less compressible, resulting in lower compression ratios, while a simple image will have a larger ratio. In our implementation, we use zlib which applies the DEFLATE algorithm to ensure lossless compression and preserve all original data \cite{zlib}.
    
    \item \textbf{Feature Count}\\
    Feature Count is determined using the Scale-Invariant Feature Transform (SIFT) algorithm to detect and count distinct features in the image \cite{lowe1999sift}. A higher Feature Count generally indicates more distinct features or greater complexity in the image. Feature Count was implemented using OpenCV's SIFT detector.
\end{itemize}

These three metrics: Image Entropy, Compression Ratio, and Feature Count, were chosen for their complementary nature in measuring visual complexity. Image Entropy measures overall information content and randomness in the pixel distributions. Compression Ratio offers insights into the redundancy of the image data, which is particularly relevant for game environments that may have repeating patterns or backgrounds. Feature Count, through the SIFT algorithm, captures the presence of distinct visual elements. Together, these metrics provide a multi-faceted view of visual complexity. In addition, combined with our existing game characteristics, these metrics form the basis for a comprehensive analysis using Random Forest regression and correlation studies.

Comparing Table \ref{tab:full-performance-comparison} and Table \ref{tab:full-game-analysis}, we can observe several interesting patterns that shed light on the relationship between game characteristics and model performance. Firstly, it's noteworthy that games with more complex action spaces, typically those with 18 distinct actions, show varied performance differences between Decision Transformer (DT) and Decision Mamba (DM). For instance, Hero and KungFuMaster, both with relatively high action space complexity (18 and 14 actions respectively), exhibit significant performance gaps but favouring DT. However, this pattern doesn't hold universally. Games like Seaquest, Alien, BankHeist, BattleZone, RoadRunner, and FishingDerby, all with 18 actions, show much smaller performance differences between the two models, with some even favouring DM. This observation suggests that action space complexity alone may not be the sole determinant of performance disparity between DT and DM. Other factors, such as visual complexity or game dynamics, likely play crucial roles in shaping model performance.

Examining the visual complexity metrics, we notice that games where DM outperforms DT, such as Breakout and Pong, tend to have lower image entropy and higher compression ratios. This could indicate that DM performs better in visually simpler environments. Conversely, games with higher image entropy and lower compression ratios, like Hero and KungFuMaster, seem to favour DT. This pattern suggests that DT might be more adept at handling visually complex environments. The feature count metric doesn't seem to have a clear correlation with performance differences. Games with both high and low feature counts show varied performance disparities between DT and DM, suggesting that this particular measure of visual complexity might not be as influential in determining model performance.

\subsubsection{Regression Analysis with Random Forest}

To quantify the relative importance of various game characteristics in determining performance differences between DT and DM, we employed a random forest regression analysis \cite{breiman2001randforest}. By using the game metrics from table \ref{tab:full-game-analysis} as input $X$and the difference in performance between DT and DM ($\text{DT norm score} - \text{DM norm score}$) as the target variable $y$. To ensure robust results given our limited sample size of 12 games, we applied a 6-fold cross validation. As shown in the pseudocode \ref{alg:random-forest}, the process starts by initializing the random forest regressor and setting up cross-validation. It then iterates through the folds, training the model on each subset and calculating the Mean Squared Error (MSE). After all folds are completed, the overall Root Mean Square Error (RMSE) and its standard deviation are calculated. Finally, the model is fitted to the entire dataset and the feature importance score is calculated. After performing the random forest regression, we get a root mean square error (RMSE) of 17.57 and a standard deviation of 12.35.

\begin{algorithm}[h!]
  % \small
  \caption{Random Forest Regression with $k$-Fold Cross-Validation}
  \begin{algorithmic}[1]
    \State \textbf{Step 1:} Initialize Random Forest and Cross-Validation
    \State \hspace{\algorithmicindent} $F \leftarrow \text{RandomForestRegressor}(n_{\text{trees}} = 1000)$
    \State \hspace{\algorithmicindent} $CV \leftarrow \text{K-Fold}(k = 6, \text{shuffle} = \text{True})$
    \State \hspace{\algorithmicindent} $S \leftarrow \emptyset$
    \State \textbf{Step 2:} Perform $k$-fold Cross-Validation
    \For{$(I_{\text{train}}, I_{\text{test}}) \in CV.\text{split}(\mathbf{X})$}
      \State $\mathbf{X}_{\text{train}}, \mathbf{X}_{\text{test}} \leftarrow \mathbf{X}[I_{\text{train}}], \mathbf{X}[I_{\text{test}}]$
      \State $\mathbf{y}_{\text{train}}, \mathbf{y}_{\text{test}} \leftarrow \mathbf{y}[I_{\text{train}}], \mathbf{y}[I_{\text{test}}]$
      \State $F.\text{fit}(\mathbf{X}_{\text{train}}, \mathbf{y}_{\text{train}})$
      \State $\hat{\mathbf{y}}_{\text{test}} \leftarrow F.\text{predict}(\mathbf{X}_{\text{test}})$
      \State $\text{MSE} \leftarrow \frac{1}{n_{\text{test}}} \sum_{i=1}^{n_{\text{test}}} (y_i - \hat{y}_i)^2$
      \State $S.\text{append}(\sqrt{\text{MSE}})$
    \EndFor
    \State \textbf{Step 3:} Calculate RMSE
    \State \hspace{\algorithmicindent} $\mu_{\text{RMSE}} \leftarrow \frac{1}{k} \sum_{s \in S} s$
    \State \hspace{\algorithmicindent} $\sigma_{\text{RMSE}} \leftarrow \sqrt{\frac{1}{k-1} \sum_{s \in S} (s - \mu_{\text{RMSE}})^2}$
    \State \textbf{Step 4:} Calculate Feature Importance
    \State \hspace{\algorithmicindent} $F.\text{fit}(\mathbf{X}, \mathbf{y})$
    \State \hspace{\algorithmicindent} $\boldsymbol{\phi} \leftarrow F.\text{feature\_importances\_}$
    \State \textbf{return} $\mu_{\text{RMSE}}, \sigma_{\text{RMSE}}, \boldsymbol{\phi}$
  \end{algorithmic}
  \label{alg:random-forest}
\end{algorithm}

\paragraph{Feature Importance}
\begin{figure}[h!]
    \centering
    \includegraphics[width=0.83\linewidth]{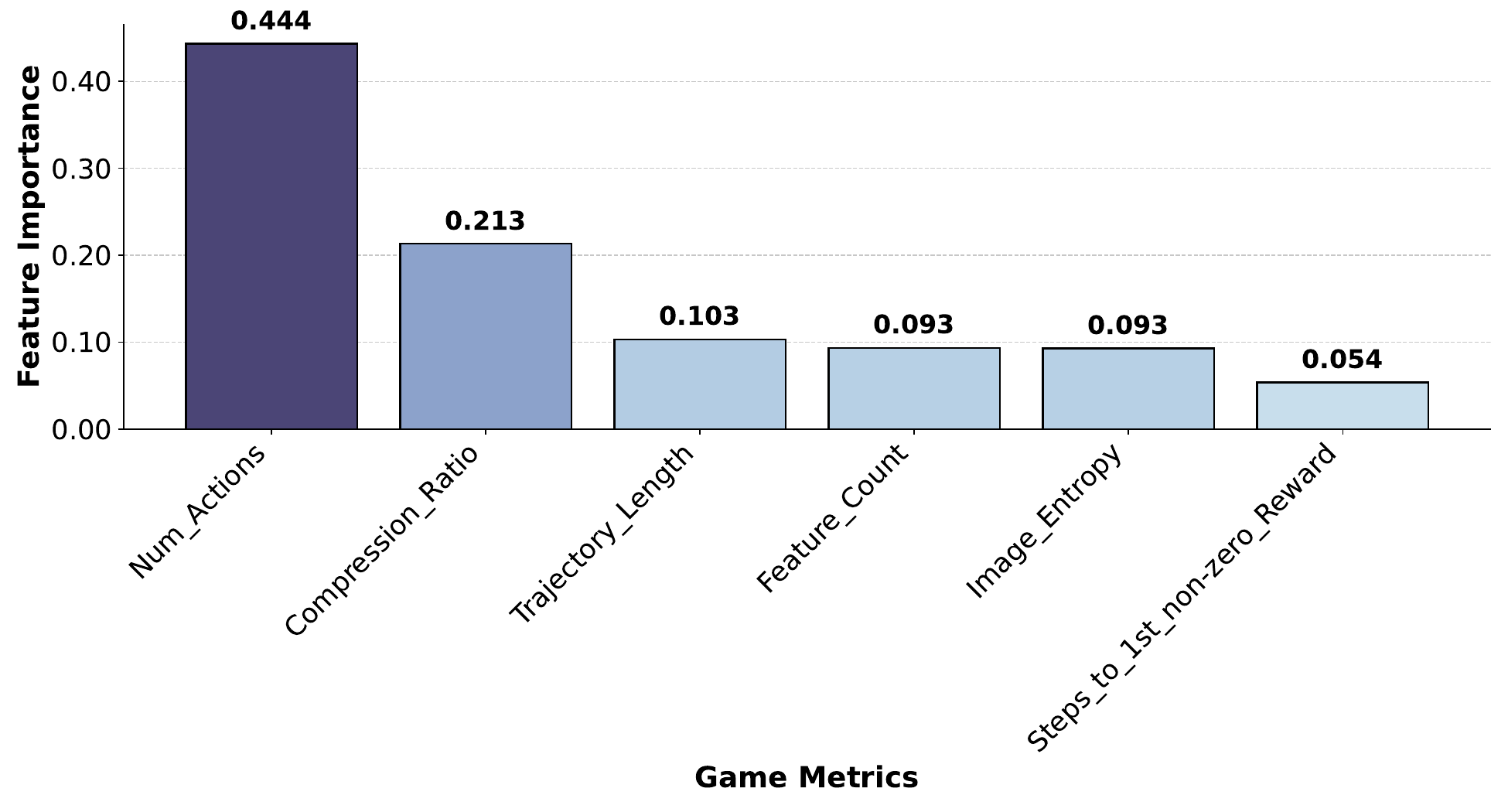}
    \captionsetup{width=0.913\linewidth}
    \caption{Feature importance of each game metrics. Higher scores indicate a greater influence on model performance difference.}
    \label{fig:feature-importance}
\end{figure}

The feature importance analysis, shown in Figure \ref{fig:feature-importance}, provides the relative influence of various game features on the performance difference between Decision Transformer (DT) and Decision Mamba (DM). The number of actions is the dominant factor, indicating that the performance difference between DT and DM tends to widen as the number of unique actions increases, generally in favour of DT. The compression ratio has the second highest significance level, followed by the average trajectory length. The feature count and the importance of image entropy rank joint fourth. The average steps to 1st non-zero reward had the weakest effect. 

These findings provide valuable insights into the factors influencing model performance. While action space complexity remains a crucial factor, the visual complexity (particularly as measured by compression ratio) of the game also plays an important role.

\paragraph{SHAP Values}

As a complementary analysis to the previous feature importance measures, we also calculated the SHAP (SHapley Additive exPlanations) values for each game metric with respect to the performance difference between Decision Transformer (DT) and Decision Mamba (DM). SHAP values provide a unified approach to explaining the output of machine learning models, potentially providing more insightful understanding of feature importance \cite{lundberg2017shap}.

\begin{figure}[h!]
    \centering
    \includegraphics[width=0.89\linewidth]{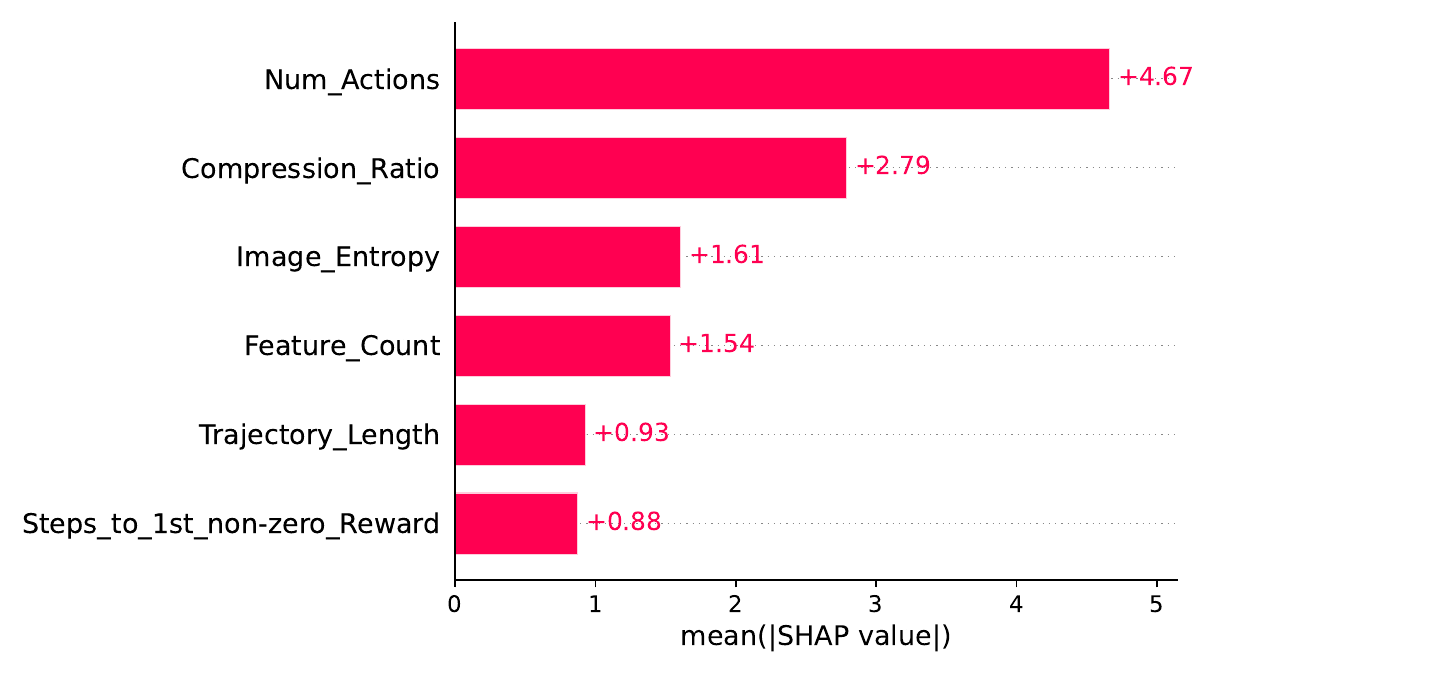}
    \captionsetup{width=0.913\linewidth}
    \caption{SHAP value feature importance. Higher values suggest a greater influence.}
    \label{fig:shap-value}
\end{figure}

Figure \ref{fig:shap-value} presents the SHAP values for each game metric. Consistent with our previous feature importance analysis, the number of actions emerges as the most influential factor, followed by the compression ratio. However, the SHAP analysis also reveals slightly different results. Image entropy and feature count still show comparable importance levels, but both with higher importance levels than previous feature importance analyses. These results further support our hypothesis that the visual complexity of games plays a significant role in the relative performance of DT and DM.

\subsubsection{Correlation Analysis}

To further investigate the relationships between game characteristics and performance differences, we conducted a correlation analysis. This analysis provides insights into the linear relationships between variables, complementing the non-linear insights gained from the Random Forest regression.

\begin{figure}[h!]
    \centering
    \includegraphics[width=0.95\linewidth]{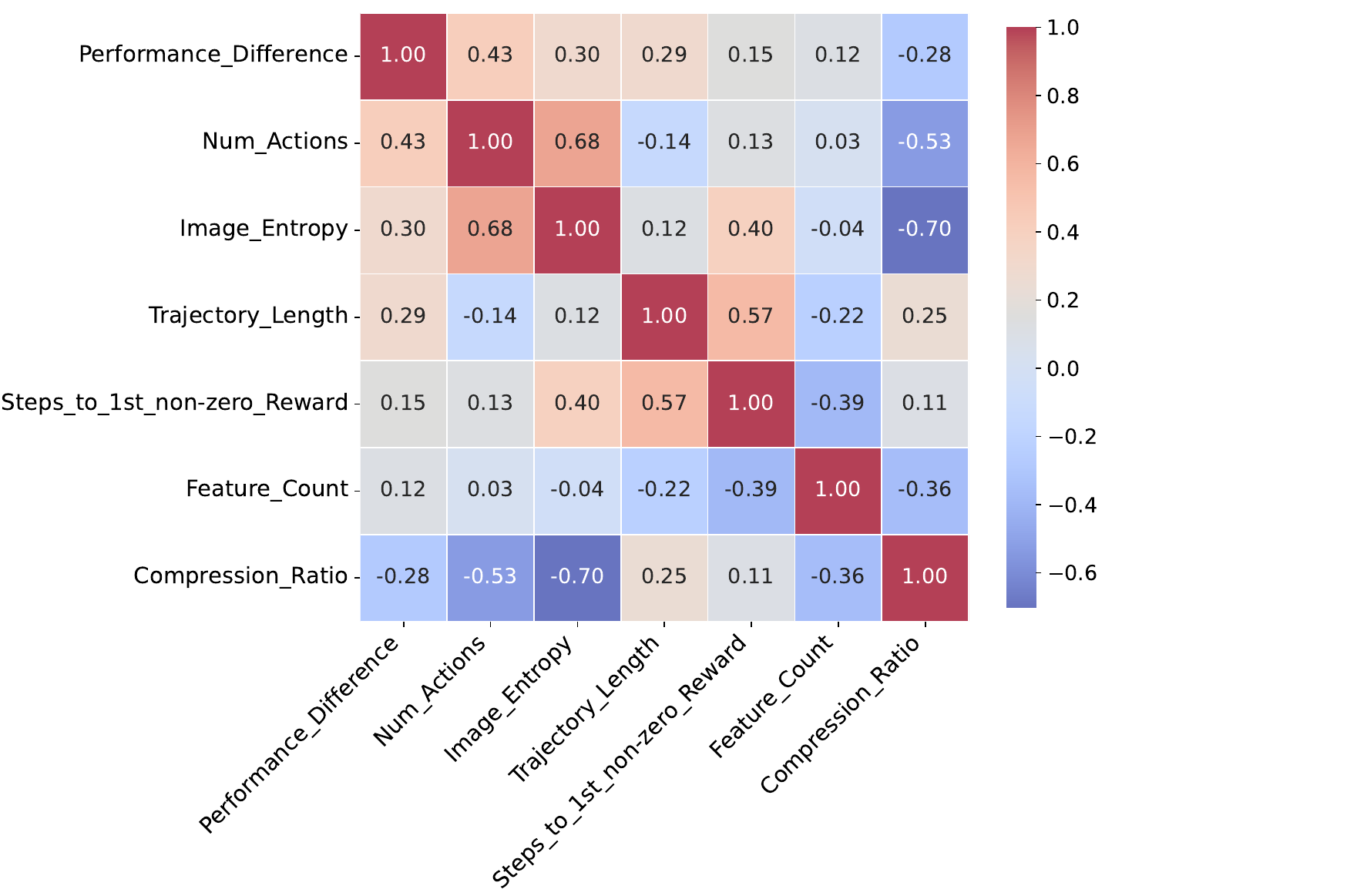}
    \captionsetup{width=0.913\linewidth}
    \caption{Correlation Matrix of Performance Difference and the Game Metrics. The colour intensity represents the strength of the correlation, with red indicating positive correlations and blue indicating negative correlations.}
    \label{fig:correlation-matrix}
\end{figure}

Similar to random forest regression, the performance difference for each game was calculated as the normalized score of Decision Transformer (DT) minus the normalized score of Decision Mamba (DM). A positive value indicates better performance by DT, while a negative value suggests DM outperformed DT. We then computed the Pearson correlation coefficient between this performance difference and each game metric, as well as between the metrics themselves \cite{berman2016pearson}. The Pearson correlation coefficient $r$ between two variables $X$ and $Y$ is defined as:

\begin{equation}
    \small
    r = \frac{\sum_{i=1}^{n} (X_i - \bar{X})(Y_i - \bar{Y})}{\sqrt{\sum_{i=1}^{n} (X_i - \bar{X})^2} \sqrt{\sum_{i=1}^{n} (Y_i - \bar{Y})^2}}
\end{equation}\\
where $X_i$ and $Y_i$ are the individual game metrics and the performance difference indexed with $i$. The coefficient $r$ ranges from -1 to 1, where -1 indicates a perfect negative linear relationship, 0 indicates no linear relationship, and 1 indicates a perfect positive linear relationship \cite{berman2016pearson}.

The resulting correlation matrix is presented in Figure \ref{fig:correlation-matrix}. In interpreting these results, we follow the guidelines where absolute values of correlation coefficients are categorized as: 0.00-0.10 ``negligible", 0.10-0.39 ``weak", 0.40-0.69 ``moderate", 0.70-0.89 ``strong", and 0.90-1.00 ``very strong" \cite{schober2018correlation}.

Examining the correlations with the performance difference, we observe several noteworthy relationships. The number of actions shows a moderate positive correlation (0.43) with the performance difference, which aligns with our earlier observation that DT tends to outperform DM in games with more complex action spaces. This correlation supports the hypothesis that action space complexity is indeed a significant factor in determining the relative performance of these two models. Image entropy exhibits a weak positive correlation (0.30) with the performance difference. This suggests that as the visual complexity of the game increases, there is a slight tendency for DT to perform better relative to DM. The compression ratio, which was identified as the second most important feature in our random forest analysis, shows a weak negative correlation (-0.28) with the performance difference. This negative correlation implies that as games become more compressible (i.e., visually simpler), there is a tendency for DM to outperform DT. This finding is consistent with our earlier observation that DM seems to excel in visually simpler environments. The average trajectory length shows weak positive correlations (0.29) with the performance difference. This suggests that DT might have a slight advantage in games with longer episodes. The average steps to the first non-zero reward and the feature count show a very weak positive correlation (0.15 and 0.12 respectively) with the performance difference, suggesting that these metrics may not significantly influence the relative performance of DT and DM.

It's also worth noting the correlations between the game metrics themselves. For instance, the number of actions shows a strong positive correlation (0.68) with image entropy, suggesting that games with more complex action spaces tend to have higher visual complexity as well.

\subsection{Analysis of the Effect of Action Space Complexity}
The Random Forest and correlation analyses provided insights into the factors influencing the performance differences between DT and DM. While these analyses highlighted the importance of both action space complexity and visual complexity, we recognized the potential for further investigation into the specific impact of action space complexity. To isolate and investigate the effect of action space complexity in these two games: Hero and Kung Fu Master, we explored a method to simplify the action space while preserving the full range of possible actions during evaluation. This led us to implement Action Fusion. 

\subsubsection{Action Fusion}
Action fusion is an approach that allows for the combination of primitive actions into fused actions, allowing the agent to perform tasks like firing and moving simultaneously in a single timestep. This method preserves the full range of possible actions during evaluation while simplifying the decision-making process during training and prediction. 

To implement action fusion, we explored two methods. The first, Simple Action Fusion, combines basic movement actions with the fire action into single composite actions. The second method, Frequency-based Action Fusion, focuses on combining actions that occur less frequently in the training data. This strategy is designed to minimize the impact on essential game mechanics while still reducing the complexity of the action space.

\begin{figure}[H]
    \centering
    \includegraphics[width=0.88\linewidth]{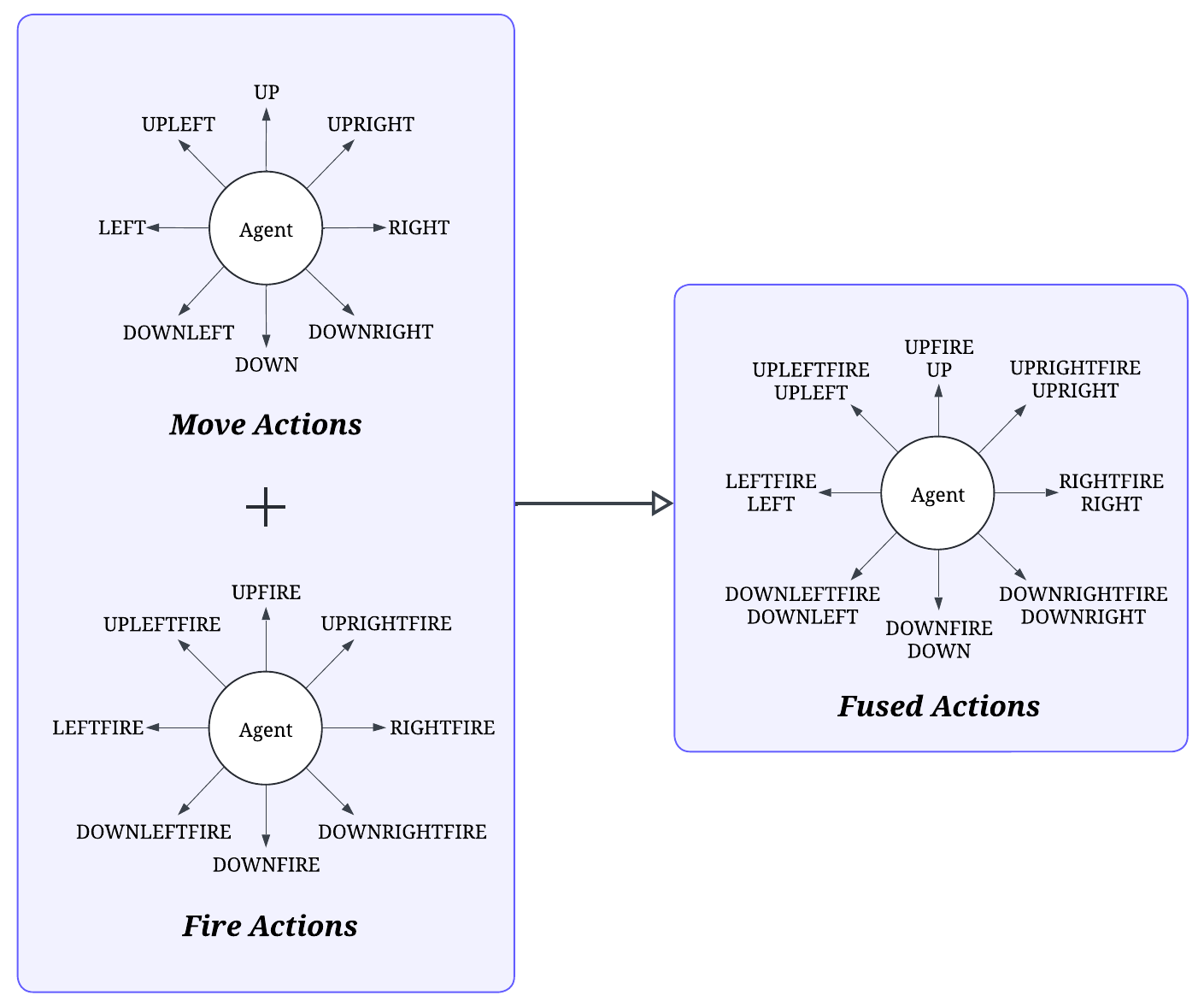}
    \caption{Simple Action Fusion: Fuse Move Actions and Fire Actions}
    \label{fig:action_fusion}
\end{figure}

\paragraph{Simple Action Fusion}
As shown in Figure \ref{fig:action_fusion}. Simple action fusion keeps NOOP and FIRE unfused, then fuse move actions and fire actions into action combinations, therefore the agent is still capable of performing the original full actions during evaluation, training and making predictions in reduced actions space. Consequently, the action space complexity for Hero is reduced from 18 to 10, and for Kung Fu Master from 14 to 9.

\begin{figure}[h!]
    \centering
    \includegraphics[width=0.95\linewidth]{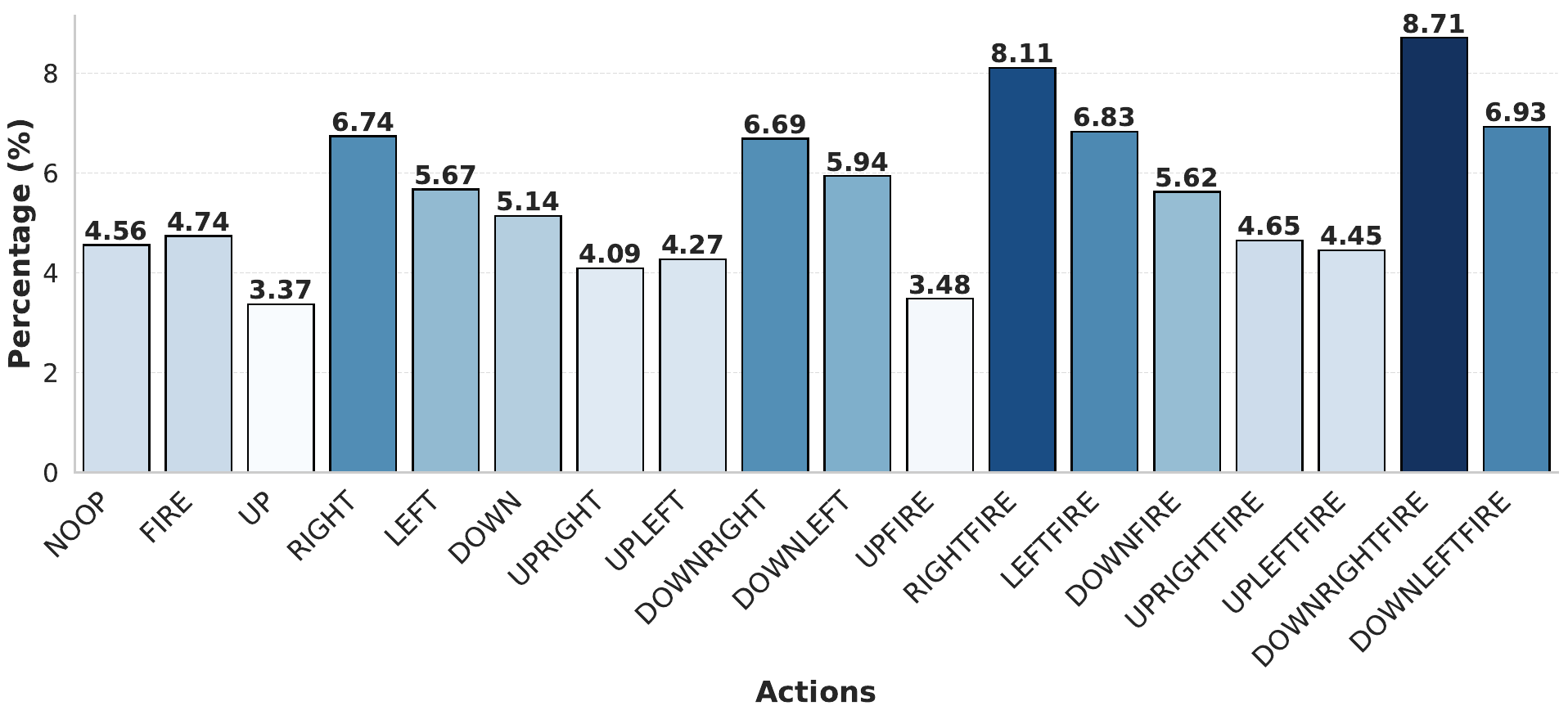}
    \caption{Action Distribution of the Last 1\% of the Dataset (Game: Hero)}
    \label{fig:action-distribution}
\end{figure}

\paragraph{Frequency-based Action Fusion}
Instead of simply fusing Move actions and Fire actions. Another feasible strategy could be fusing according to the action distributions of the last 1\% of the dataset, which is considered to be the expert knowledge of the game \cite{agarwal2020dqn}. The relatively non-frequent actions are fused, for example, as shown in Figure \ref{fig:action-distribution}, UP(3.37\%) and UPFIRE(3.48\%) will be fused into (UPFIRE, UP). And UPRIGHT(4.09\%) and UPLEFT(4.27\%) will be fused into (UPRIGHT, UPLEFT). Since we fuse non-frequent actions, this method is expected to reduce the effect on the game caused by action fusion.

Implementing action fusion (Simple or Frequency-based) required modifications to the Arcade Learning Environment (ALE) to allow the agent(or the model) to perform fused actions. ALE serves as the foundational backend for OpenAI Gym's Atari environments and allows for more fine-grained control over the emulation process. The implementation process involved several key steps. First, we created the \textit{Env} class in \textit{trainer\_atari.py}, which serves as a wrapper around ALE. Next, we created a mapping between the original action space and the fused action space. This mapping was implemented as a dictionary, \textit{fused\_action\_map}, which associates each fused action with a list of primitive actions. The mapping strategy varies depending on the specific game; detailed strategies are provided in Appendix \ref{tab:nonfrequent-action-fusion}. Overall, this implementation preserves the full action space during evaluation. When the trained model interacts with the game environment, it still has access to all original actions, but it makes decisions in the simplified action space.

\subsubsection{Results of Action Fusion}
Table \ref{tab:original-results} and Table \ref{tab:action-fusion} present the original results and the outcomes of both action fusion methods, respectively. To directly compare the original and action-fused results, we maintain consistent values (Random Walk scores and Human Players soirees) when calculating normalized scores. Although action fusion simplifies the game's action space, it preserves the core game mechanics and objectives. The fused actions still enable the agent to perform all original game actions, with the reward assignment mechanism controlled by ALE remaining unchanged. While this method may introduce slight bias, it ensures comparability across experiments. 

\begin{table}[h!]
\centering
\small
\begin{tabular}{@{}lcc@{}}
\toprule
Game & DT & DM \\
\midrule
Hero & \textbf{30.37 $\pm$ 4.47} & 7.77 $\pm$ 0.99 \\
KungFuMaster & \textbf{29.41 $\pm$ 6.48} & 5.29 $\pm$ 0.89 \\
\bottomrule
\end{tabular}
\captionsetup{width=0.87\linewidth}
\caption{Original Results without Action Fusion. Context length 10. Scores are averaged over 3 runs, each with a unique random seed. The better scores are \textbf{bold}.}
\label{tab:original-results}
\end{table}

\begin{table}[h!]
\centering
\small
\begin{tabular}{@{}lcccc@{}}
\toprule
\multirow{2}{*}{Game} & \multicolumn{2}{c}{Simple Action Fusion} & \multicolumn{2}{c}{Frequency-based Action Fusion} \\
\cmidrule(lr){2-3} \cmidrule(lr){4-5}
& DT & DM & DT & DM \\
\midrule
Hero & \textbf{18.72 $\pm$ 3.67} & 7.07 $\pm$ 0.73 & \textbf{16.06 $\pm$ 0.74} & 7.26 $\pm$ 1.21 \\
KungFuMaster & 3.09 $\pm$ 3.19 & \textbf{3.64 $\pm$ 0.95} & 1.55 $\pm$ 0.86 & \textbf{2.79 $\pm$ 0.33} \\
\bottomrule
\end{tabular}
\captionsetup{width=0.87\linewidth}
\caption{Action Fusion Results: Normalized scores (mean $\pm$ std) for Simple Action Fusion and Frequency-based Action Fusion across games Hero and KungFuMaster, with context length 10. Scores are averaged over 3 separate runs, each with a unique random seed. The better scores are \textbf{bold}.}
\label{tab:action-fusion}
\end{table}

In the game Hero, the normalized score of Decision Transformer dropped from the original 30.37 to 18.72 for Simple Action Fusion and 16.06 for Frequency-based Action Fusion, while the Decision Mamba's performance in Hero remained relatively stable, with only slight variations across different fusion strategies. For KungFuMaster, both DT and DM experienced significant performance drops with action fusion, but the relative performance between the two models shifted. In the original setup, DT substantially outperformed DM (29.41 vs 5.29). However, with action fusion applied, DM slightly edged out DT in both fusion strategies.

The results suggest that simplifying the action space, even when preserving the full range of actions during evaluation, impacts the models' ability to learn optimal strategies. The performance drops observed with both fusion methods, particularly for DT, indicate that action space complexity is indeed a significant factor in model performance. However, the persistent performance gaps between DT and Decision Mamba (DM), especially in games like Hero, suggest that action space complexity alone cannot fully account for the disparities. This finding points to the need for more investigation into other potential factors influencing performance. Given that both models process game frames as input, visual complexity emerges as a promising avenue for further exploration. Consequently, we propose a new hypothesis: the performance gap between DT and DM in games like Hero and KungFuMaster is likely influenced by multiple factors, with action space complexity being just one component of a more complex interplay of game characteristics.

\section{Future Work}
The above analysis extends our understanding of the relationship between game characteristics and performance differences in Decision Transformer (DT) and Decision Mamba (DM). The analysis confirms the importance of action space complexity, it also reveals the key role of visual complexity in it. These findings emphasise the need for further research. One potential direction is to investigate the fundamental mechanisms in visual processing capabilities. This may need a theoretical analysis of attention matrices in DT and state space representations in DM of games with different visual complexities. Moreover, future research could explore hybrid architectures that combine the advantages of DT and DM, potentially yielding more robust performance in a wider range of environments.

\section{Conclusion}
In conclusion, this study has conducted a comprehensive evaluation of Decision Transformer (DT) and Decision Mamba (DM) across selected 12 Atari games. This research started by noting the differences in performance between DT and DM in the games Hero and Kung Fu Master, leading to a systematic analysis of the game characteristics influencing these discrepancies. The random forest regression analysis highlights the importance of action space complexity, as well as visual complexity, by calculating the feature importance and SHAP values. Our correlation analysis also supports these findings, showing a moderate positive correlation between performance differences and game characteristics such as the number of actions and some weak correlation with image entropy. In addition, to further investigate the effect of action space complexity, we implemented action fusion, including simple action fusion and frequency-based action fusion, further highlighting the important role of action space complexity. Overall, our analysis suggests that action space complexity and visual complexity are the main factors that influence the performance gap between Decision Transformer and Decision Mamba. DM tended to perform better with simpler action space and visual environments, and DT showed an advantage in games with complicated actions and visual components. For future research, it can focus on theoretical analysis of the attention patterns in DT and the state space representations in DM. Alternatively, researchers can focus on developing hybrid models that combine the advantages of DT and DM to improve the model's ability to handle complex visual elements.

\bibliographystyle{IEEEtran}
\bibliography{references}

\newpage
\begin{appendices}
\section{Hyperparameters}
\label{sec:hyper}

\begin{table}[H]
\centering
\small
\begin{tabular}{ll}
\toprule
\textbf{Hyperparameter}          & \textbf{Value}   \\
\midrule
Number of layers                             & 6                                             \\
Embedding dimension                          & 128                                           \\
Batch size                                   & 256                                           \\
Context length K                             & 10, 30, 50                                    \\
Max epochs                                   & 5                                             \\
Dropout                                      & 0.1                                           \\
Learning rate                                & 0.0006                                        \\
Adam betas                                   & (0.9, 0.95)                                   \\
Grad norm clip                               & 1.0                                           \\
Weight decay                                 & 0.1                                           \\
Learning rate decay                          & Linear warmup and cosine decay                \\
Initial expected return (during evaluation)  & 5*max game score appeared in the dataset \\
3 Random seeds (default)                     & 123, 132, 231                                 \\
5 Random seeds (initial experiments)         & 123, 132, 231, 231, 312              \\
\bottomrule
\end{tabular}
\caption{Hyperparameter settings}
\label{tab:hyperparameters}
\end{table}

\section{Examples of Game States (four consecutive frames)}
\label{sec:game-states}

\begin{figure}[H]
    \centering
    \includegraphics[width=0.9\linewidth]{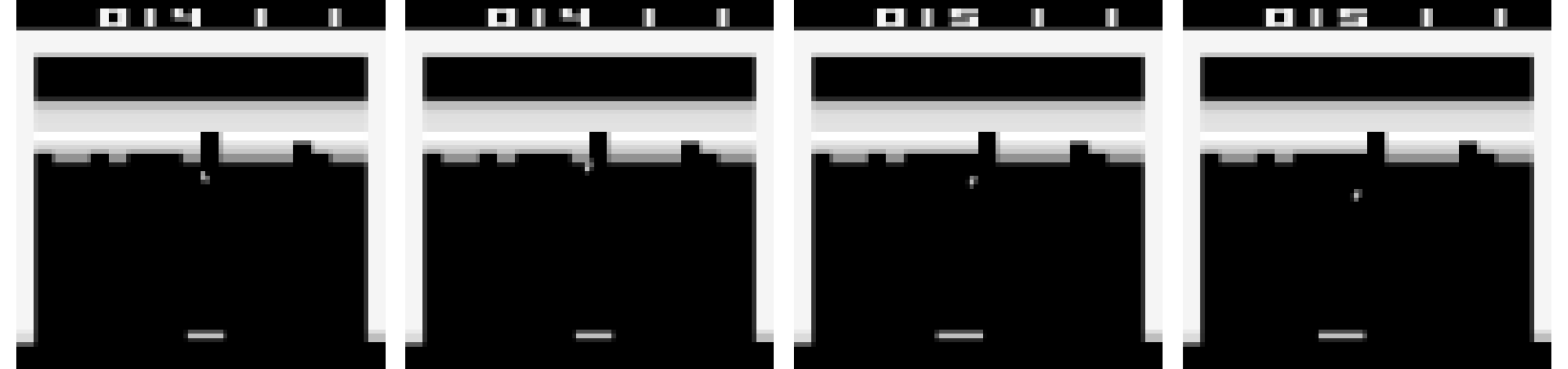}
    \caption{Breakout}
    \label{fig:breakout}
\end{figure}
\vspace{-20pt}
\begin{figure}[H]
    \centering
    \includegraphics[width=0.9\linewidth]{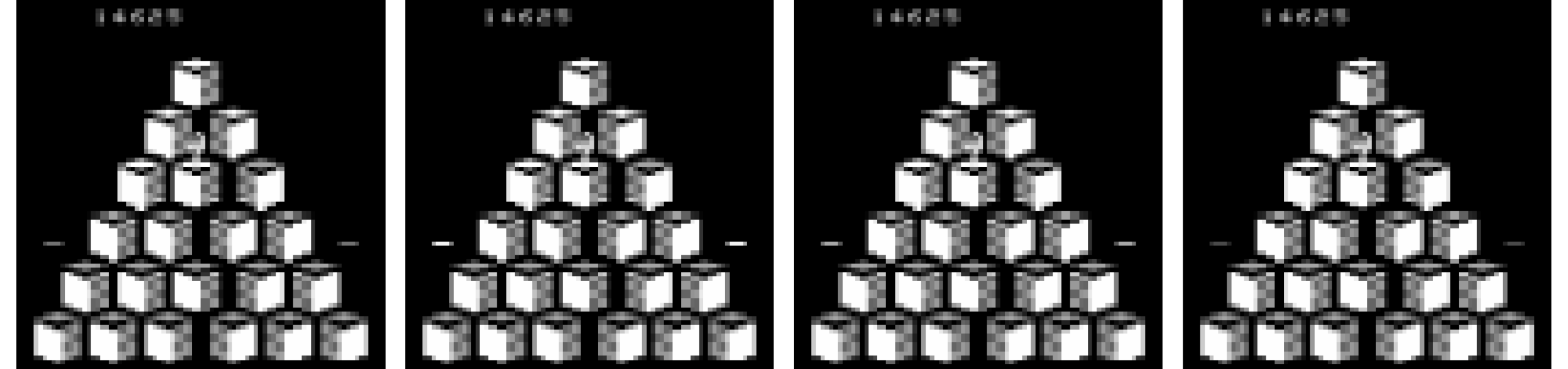}
    \caption{Qbert}
    \label{fig:qbert}
\end{figure}
\vspace{-20pt}
\begin{figure}[H]
    \centering
    \includegraphics[width=0.9\linewidth]{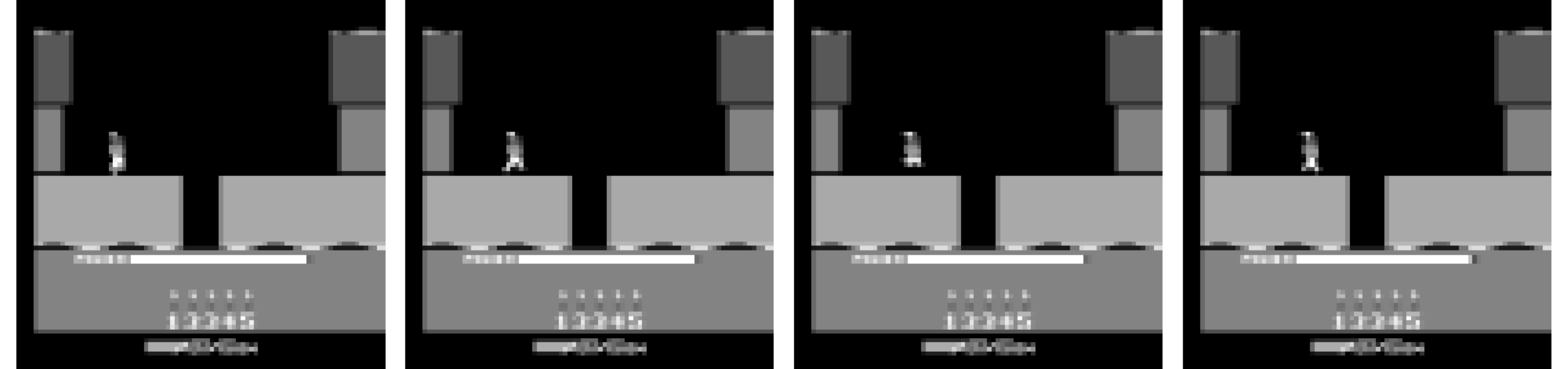}
    \caption{Hero}
    \label{fig:hero}
\end{figure}
\vspace{-20pt}
\begin{figure}[H]
    \centering
    \includegraphics[width=0.9\linewidth]{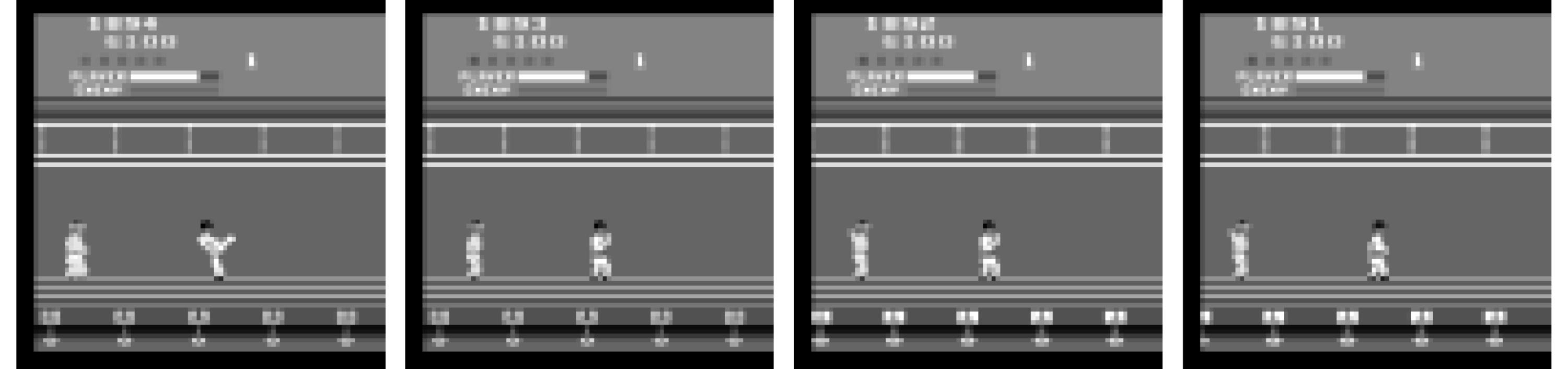}
    \caption{Kung Fu Master}
    \label{fig:kfm}
\end{figure}

\section{Max Return and Expected Return}
\begin{table}[H]
\centering
\small
\begin{tabular}{@{}lcc@{}}
\toprule
\textbf{Games}          & \textbf{Max Return} & \textbf{Expected Return (5*max)} \\ 
\midrule
Breakout                & 104                 & 520                              \\
Qbert                   & 640                 & 3200                             \\
Hero                    & 190                 & 950                              \\
KungFuMaster            & 284                 & 1420                             \\
Pong                    & 21                  & 105                              \\
Seaquest                & 314                 & 1570                             \\
Alien                   & 213                 & 1065                             \\
BankHeist               & 136                 & 680                              \\
BattleZone              & 32                  & 160                              \\
RoadRunner              & 270                 & 1350                             \\
FishingDerby            & 61                  & 305                              \\
SpaceInvaders           & 288                 & 1440                             \\
\bottomrule
\end{tabular}
\caption{Max Return and Expected Return (5 times max) for selected games}
\end{table}

\section{Game scores for Random Walk and Human Player (Full table)}
\label{sec:scores}

\begin{table}[H]
\centering
\small
\begin{tabular}{@{}lcc@{}}
\toprule
\textbf{Game} & \textbf{Random Walk} & \textbf{Human Player} \\
\midrule
Breakout & 1.7 & 30.5 \\
Qbert & 163.9 & 13455.0 \\
Hero & 1027.0 & 30826.4 \\
KungFuMaster & 258.5 & 22736.3 \\
Pong & -20.7 & 14.6 \\
Seaquest & 68.4 & 42054.7 \\
Alien & 227.8 & 7127.7 \\
BankHeist & 14.0 & 753.0 \\
BattleZone & 2360.0 & 37187.5 \\
RoadRunner & 11.5 & 7845.0 \\
FishingDerby & -92.0 & -39.0 \\
SpaceInvaders & 148.0 & 1669.0 \\
\bottomrule
\end{tabular}
\caption{Game scores by Random Walk and Human Players, updated to reflect more precise values \cite{hafner2021master, ye2021master}.}
\label{tab:updated-rand-human-score}
\end{table}

\section{Normalized scores without removing outliers}
\label{sec:raw-norm-score}

\begin{table}[H]
\centering
\small      
\begin{tabularx}{\textwidth}{@{}X *{6}{>{\raggedleft\arraybackslash}X}@{}}
\toprule
               & \multicolumn{3}{c}{Decision Transformer (DT)} & \multicolumn{3}{c}{Decision Mamba (DM)} \\
               \cmidrule(lr){2-4} \cmidrule(lr){5-7}
Game           & 10                    & 30                    & 50                    & 10                    & 30                    & 50                    \\ \midrule
Breakout       & 267.36    & 218.75    & 208.33    & \textbf{343.12}    & \textbf{384.79}    & \textbf{340.35}   \\[-1ex]
               & $\pm$ 90.69    & $\pm$ 97.20    & $\pm$ 51.74    & \textbf{$\pm$ 65.35}    & \textbf{$\pm$ 54.29}    & \textbf{$\pm$ 93.75}   \\
\midrule
Qbert          & \textbf{29.67}     & 9.06     & 7.89      & 26.54      & \textbf{23.05}     & \textbf{22.59}      \\[-1ex]
               & \textbf{$\pm$ 12.84}     & $\pm$ 9.40     & $\pm$ 6.58      & $\pm$ 1.01      & \textbf{$\pm$ 7.74}     & \textbf{$\pm$ 1.84}      \\
\midrule
Hero           & \textbf{31.62}      & \textbf{28.05}      & \textbf{25.54}      & 7.70       & 6.95       & 7.69       \\[-1ex]
               & \textbf{$\pm$ 4.16}      & \textbf{$\pm$ 4.44}      & \textbf{$\pm$ 8.32}      & $\pm$ 0.96       & $\pm$ 0.43       & $\pm$ 1.42       \\
\midrule
KungFuMaster   & \textbf{19.65}      & \textbf{11.24}      & \textbf{9.68}     & 6.32       & 6.72       & 6.03       \\[-1ex]
               & \textbf{$\pm$ 14.15}      & \textbf{$\pm$ 5.84}      & \textbf{$\pm$ 8.01}     & $\pm$ 1.75       & $\pm$ 1.46       & $\pm$ 2.80       \\
\bottomrule
\end{tabularx}
\caption{Normalized score (mean $\pm$ std) of Decision Mamba and Decision Transformer across four games with context lengths of 10, 30, and 50. Each game was run with 5 random seeds per context length. \textbf{Bold} indicates the better mean performance.}
\label{tab:without-outliers}
\end{table}

\section{Action Distribution of the last 1\% of the dataset}
\label{sec:act-dist}

\begin{table}[H]
\centering
\footnotesize
\begin{tabular}{|c|l|c|}
\hline
\textbf{Index} & \textbf{Action}        & \textbf{Percentage} \\ \hline
0              & NOOP                   & 4.56\%              \\ \hline
1              & FIRE                   & 4.74\%              \\ \hline
2              & UP                     & 3.37\%              \\ \hline
3              & RIGHT                  & 6.74\%              \\ \hline
4              & LEFT                   & 5.67\%              \\ \hline
5              & DOWN                   & 5.14\%              \\ \hline
6              & UPRIGHT                & 4.09\%              \\ \hline
7              & UPLEFT                 & 4.27\%              \\ \hline
8              & DOWNRIGHT              & 6.69\%              \\ \hline
9              & DOWNLEFT               & 5.94\%              \\ \hline
10             & UPFIRE                 & 3.48\%              \\ \hline
11             & RIGHTFIRE              & 8.11\%              \\ \hline
12             & LEFTFIRE               & 6.83\%              \\ \hline
13             & DOWNFIRE               & 5.62\%              \\ \hline
14             & UPRIGHTFIRE            & 4.65\%              \\ \hline
15             & UPLEFTFIRE             & 4.45\%              \\ \hline
16             & DOWNRIGHTFIRE          & 8.71\%              \\ \hline
17             & DOWNLEFTFIRE           & 6.93\%              \\ \hline
\end{tabular}
\caption{Distribution of Actions for Hero}
\end{table}

\begin{table}[H]
\centering
\footnotesize
\begin{tabular}{|c|l|c|}
\hline
\textbf{Index} & \textbf{Action}        & \textbf{Percentage} \\ \hline
0              & NOOP                   & 6.43\%              \\ \hline
1              & UP                     & 6.16\%              \\ \hline
2              & RIGHT                  & 4.79\%              \\ \hline
3              & LEFT                   & 7.96\%              \\ \hline
4              & DOWN                   & 6.70\%              \\ \hline
5              & DOWNRIGHT              & 7.89\%              \\ \hline
6              & DOWNLEFT               & 7.14\%              \\ \hline
7              & RIGHTFIRE              & 7.54\%              \\ \hline
8              & LEFTFIRE               & 8.71\%              \\ \hline
9              & DOWNFIRE               & 7.15\%              \\ \hline
10             & UPRIGHTFIRE            & 6.70\%              \\ \hline
11             & UPLEFTFIRE             & 8.12\%              \\ \hline
12             & DOWNRIGHTFIRE          & 7.10\%              \\ \hline
13             & DOWNLEFTFIRE           & 7.60\%              \\ \hline
\end{tabular}
\caption{Distribution of Actions for KungFuMaster}
\end{table}

\section{Action Fusion Strategy}
\label{sec:act-fuse}

\subsection{Simple Action Fusion}
\begin{table}[H]
\centering
\footnotesize
\begin{tabular}{|l|l|l|}
\hline
\textbf{Game} & \textbf{Original Actions} & \textbf{Fused Actions} \\
\hline
\multirow{18}{*}{Hero} & 0: NOOP & 0: NOOP \\
& 1: FIRE & 1: FIRE \\
& 2: UP & 2: [UP, UPFIRE] \\
& 3: RIGHT & 3: [RIGHT, RIGHTFIRE] \\
& 4: LEFT & 4: [LEFT, LEFTFIRE] \\
& 5: DOWN & 5: [DOWN, DOWNFIRE] \\
& 6: UPRIGHT & 6: [UPRIGHT, UPRIGHTFIRE] \\
& 7: UPLEFT & 7: [UPLEFT, UPLEFTFIRE] \\
& 8: DOWNRIGHT & 8: [DOWNRIGHT, DOWNRIGHTFIRE] \\
& 9: DOWNLEFT & 9: [DOWNLEFT, DOWNLEFTFIRE] \\
& 10: UPFIRE & \\
& 11: RIGHTFIRE & \\
& 12: LEFTFIRE & \\
& 13: DOWNFIRE & \\
& 14: UPRIGHTFIRE & \\
& 15: UPLEFTFIRE & \\
& 16: DOWNRIGHTFIRE & \\
& 17: DOWNLEFTFIRE & \\
\hline
\multirow{14}{*}{KungFuMaster} & 0: NOOP & 0: NOOP \\
& 1: UP & 1: UP \\
& 2: RIGHT & 2: [RIGHT, RIGHTFIRE] \\
& 3: LEFT & 3: [LEFT, LEFTFIRE] \\
& 4: DOWN & 4: [DOWN, DOWNFIRE] \\
& 5: DOWNRIGHT & 5: [DOWNRIGHT, DOWNRIGHTFIRE] \\
& 6: DOWNLEFT & 6: [DOWNLEFT, DOWNLEFTFIRE] \\
& 7: RIGHTFIRE & 7: UPRIGHTFIRE \\
& 8: LEFTFIRE & 8: UPLEFTFIRE \\
& 9: DOWNFIRE & \\
& 10: UPRIGHTFIRE & \\
& 11: UPLEFTFIRE & \\
& 12: DOWNRIGHTFIRE & \\
& 13: DOWNLEFTFIRE & \\
\hline
\end{tabular}
\caption{Simple Action Fusion Strategy}
\label{tab:simple-action-fusion}
\end{table}

\subsection{Frequency-based Action Fusion}
\begin{table}[H]
\centering
\footnotesize
\begin{tabular}{|l|l|l|}
\hline
\textbf{Game} & \textbf{Original Actions} & \textbf{Fused Actions} \\
\hline
\multirow{18}{*}{Hero} & 0: NOOP & 0: [UP, UPFIRE] \\
& 1: FIRE & 1: [FIRE, UPLEFT, UPLEFTFIRE] \\
& 2: UP & 2: [NOOP] \\
& 3: RIGHT & 3: [RIGHT, DOWNRIGHT] \\
& 4: LEFT & 4: [LEFT, LEFTFIRE] \\
& 5: DOWN & 5: [DOWN, DOWNFIRE] \\
& 6: UPRIGHT & 6: [UPRIGHT, UPRIGHTFIRE] \\
& 7: UPLEFT & 7: [RIGHTFIRE, DOWNRIGHTFIRE] \\
& 8: DOWNRIGHT & 8: [DOWNLEFT, DOWNLEFTFIRE] \\
& 9: DOWNLEFT & 9: [UPLEFT, UPLEFTFIRE] \\
& 10: UPFIRE & \\
& 11: RIGHTFIRE & \\
& 12: LEFTFIRE & \\
& 13: DOWNFIRE & \\
& 14: UPRIGHTFIRE & \\
& 15: UPLEFTFIRE & \\
& 16: DOWNRIGHTFIRE & \\
& 17: DOWNLEFTFIRE & \\
\hline
\multirow{14}{*}{KungFuMaster} & 0: NOOP & 0: [UP, NOOP] \\
& 1: UP & 1: [RIGHT, UPRIGHTFIRE] \\
& 2: RIGHT & 2: [DOWN, RIGHTFIRE] \\
& 3: LEFT & 3: [LEFT, UPLEFTFIRE] \\
& 4: DOWN & 4: [DOWNFIRE, LEFTFIRE] \\
& 5: DOWNRIGHT & 5: [DOWNRIGHT, DOWNRIGHTFIRE] \\
& 6: DOWNLEFT & 6: [DOWNLEFT, DOWNLEFTFIRE] \\
& 7: RIGHTFIRE & 7: [UPLEFTFIRE, LEFTFIRE] \\
& 8: LEFTFIRE & \\
& 9: DOWNFIRE & \\
& 10: UPRIGHTFIRE & \\
& 11: UPLEFTFIRE & \\
& 12: DOWNRIGHTFIRE & \\
& 13: DOWNLEFTFIRE & \\
\hline
\end{tabular}
\caption{Frequency-based Action Fusion Strategy}
\label{tab:nonfrequent-action-fusion}
\end{table}

\end{appendices}

\end{document}